\documentclass[conference]{IEEEtran}
\IEEEoverridecommandlockouts
\usepackage{cite}
\usepackage{amsmath,amssymb,amsfonts,nccmath}
\usepackage{graphicx}
\usepackage{textcomp}
\usepackage{xcolor}

\usepackage{bm}
\usepackage[noend]{algpseudocode}
\usepackage{algorithmicx}
\usepackage{algorithm}
\usepackage{multirow}
\usepackage{booktabs}
\usepackage{pifont}
\usepackage{verbatim}
\usepackage{balance}
\usepackage{footmisc}

\def\BibTeX{{\rm B\kern-.05em{\sc i\kern-.025em b}\kern-.08em
    T\kern-.1667em\lower.7ex\hbox{E}\kern-.125emX}}
\renewcommand{\baselinestretch}{0.97}
\begin{document}

\title{Astraea: Self-balancing Federated Learning for Improving Classification Accuracy of Mobile Deep Learning Applications}

\author{\IEEEauthorblockN{Moming Duan\IEEEauthorrefmark{1}, 
		Duo Liu\IEEEauthorrefmark{1},
	Xianzhang Chen\IEEEauthorrefmark{1},
Yujuan Tan\IEEEauthorrefmark{1},
Jinting Ren\IEEEauthorrefmark{1},
Lei Qiao\IEEEauthorrefmark{2},
Liang Liang\IEEEauthorrefmark{1}}
\IEEEauthorblockA{\IEEEauthorrefmark{1}College of Computer Science, Chongqing University, Chongqing, China\\
}
\IEEEauthorblockA{\IEEEauthorrefmark{2}Beijing Institute of Control Engineering, Beijing, China\\
}
}

\maketitle

\setlength\footnotemargin{0em}
\let\thefootnote\relax\footnotetext{ \rule[0.25\baselineskip]{0.5\columnwidth}{0.5pt}\\
Corresponding authors: Duo Liu and Yujuan Tan, College of Computer Science, Chongqing University, Chongqing, China.\\
E-mail: \{liuduo, tanyujuan\}@cqu.edu.cn.}

\begin{abstract}
Federated learning (FL) is a distributed deep learning method which enables multiple participants, such as mobile phones and IoT devices, to contribute a neural network model while their private training data remains in local devices. This distributed approach is promising in the edge computing system where have a large corpus of decentralized data and require high privacy. However, unlike the common training dataset, the data distribution of the edge computing system is imbalanced which will introduce biases in the model training and cause a decrease in accuracy of federated learning applications.
In this paper, we demonstrate that the imbalanced distributed training data will cause accuracy degradation in FL. To counter this problem, we build a self-balancing federated learning framework call Astraea, which alleviates the imbalances by 1) Global data distribution based data augmentation, and 2) Mediator based multi-client rescheduling. The proposed framework relieves global imbalance by runtime data augmentation, and for averaging the local imbalance, it creates the mediator to reschedule the training of clients based on Kullback–Leibler divergence (KLD) of their data distribution. Compared with \textit{FedAvg}, the state-of-the-art FL algorithm, Astraea shows +5.59\% and +5.89\% improvement of top-1 accuracy on the imbalanced EMNIST and imbalanced CINIC-10 datasets, respectively. Meanwhile, the communication traffic of Astraea can be 92\% lower than that of \textit{FedAvg}.
\end{abstract}


\section{Introduction}
Federated Learning (FL) is a promising distributed neural network training approach for deep learning applications such as image classification~\cite{krizhevsky2012imagenet} and nature language process~\cite{hinton2012deep}.
FL enables the mobile devices to collaboratively train a shared neural network model with the training data distributed on the local devices.
In a FL application, any mobile device can participate in the neural network model training task as a client. 
Each client independently trains the neural network model based on its local data.
A FL server 
then averages the models' updates from a random subset of FL clients and aggregates them into a new global model. 
In this way, FL not only ensures privacy, costs lower latency, but also makes the mobile applications adaptive to the changes of its data.

Nevertheless, a main challenge of the mobile federated learning is that the training data is unevenly distributed on the mobile devices, which results in low prediction accuracy.
Several efforts have been made to tackle the challenge. 
McMahan \textit{et al.} propose a communication-efficient FL algorithm \textit{Federated Averaging (FedAvg)}~\cite{mcmahan2016communication}, and show that the CNN model trained by FedAvg can achieve 99\% test accuracy on non-IID MNIST dataset, i.e., any particular user of the local dataset is not representative of the population distribution.
Zhao \textit{et al.}~\cite{zhao2018federated}  point out that the CNN model trained by \textit{FedAvg} on non-IID CIFAR-10 dataset has 37\% accuracy loss. 
Existing studies assume that the expectation of the global data distribution is balanced even though the volume of data on the devices may be disproportionate. In most real scenarios of distributed mobile devices, however, the global data distribution is imbalanced.

In this paper, we consider one more type of imbalanced distribution, named global imbalanced, of distributed training data.
In global imbalanced distribution, the collection of distributed data is class imbalanced.  
We draw a global imbalanced subset from EMNIST dataset and explore its impact on the accuracy of FL in Section \ref{sec:motivation}. 
The experimental results show that the global imbalanced training data leads to 7.92\% accuracy loss for \textit{FedAvg}.

The accuracy degradation caused by imbalances drives us to design a novel self-balancing federated learning framework, called Astraea. 
The Astraea framework counterweighs the training of FL with imbalanced datasets by two strategies.
First, before training the model, Astraea performs data augmentation~\cite{wong2016understanding} to alleviate global imbalance.
Second, Astraea proposes to use some mediators to reschedule the training of clients according to the KLD between the mediators and the uniform distribution. 
By combining the training of skewed clients, the mediators may be able to achieve a new partial equilibrium.

With the above methods, Astraea improves 5.59\% top-1 accuracy on the imbalanced EMNIST and 5.89\% on imbalanced CINIC-10~\cite{darlow2018cinic} over \textit{FedAvg}. 
Our rescheduling strategy can significantly reduce the impact of local imbalance and decrease the mean of the KLD between the mediators and the uniform distribution to below 0.2. 
The proposed framework is also communication-efficient. 
For example, the experimental results show that Astraea can reduce 92\% communication traffic than that of \textit{FedAvg} in achieving 75\% accuracy on imbalanced EMNIST.

The main contributions of this paper are summarized as follows.
\begin{itemize}
	\item We first find out that the global imbalanced training data will degrade the accuracy of CNN models trained by FL.
	
	\item We propose a self-balancing federated learning framework, Astraea, along with two strategies to prevent the bias of training caused by imbalanced data distribution. 
	
	\item We implement and measure the proposed Astraea based on the Tensorflow Federated Framework~\cite{abadi2016tensorflow}. The experimental results show that Astraea can efficiently retrieve 70.5\% accuracy loss on imbalanced EMNIST and retrieve 47.83\% accuracy loss on imbalanced CINIC-10 dataset.
\end{itemize}

\section{Background and Motivation}

\subsection{Background}
\textbf{Federated learning}. FL is proposed in~\cite{mcmahan2016communication}, which includes the model aggregation algorithm \textit{FedAvg}. In the FL system, all clients calculate and update their weights using asynchronous stochastic gradient descent (SGD) in parallel, 
then a server collects the updates of clients and aggregates them using \textit{FedAvg} algorithm. With the distributed training method, a number of mobile deep learning applications based on FL have recently emerged. 
Hard \textit{et al.}~\cite{hard2018federated} improve the next word predictions of Google keyboard through FL.
Bonawitz \textit{et al.}~\cite{bonawitz2019towards} build a large scale FL system in the domain of mobile devices. 

Recent research on federated learning has focused on reducing communication overhead~\cite{konevcny2016federated,mcmahan2016communication,samarakoon2018federated, liu2019fitcnn} and protecting privacy~\cite{bonawitz2017practical,agarwal2018cpsgd,wang2019beyond,xu2019verifynet}, but only a few studies have noticed the problem of accuracy degradation due to imbalance~\cite{zhao2018federated,sattler2019robust}. However,~\cite{zhao2018federated,sattler2019robust,mcmahan2016communication} only discuss the impact of local imbalanced data and assume the global data distribution is balanced, which is rare in the distributed mobile system.

\textbf{Imbalanced data learning}. Most real-world classification tasks have class imbalance which will increase the bias of machine learning algorithms. Learning with imbalanced distribution is a classic problem in the field of data science~\cite{he2008learning}, and its main solutions are sampling and ensemble learning. Undersampling method samples the dataset to get a balanced subset, which is easy to implement. This method requires a large data set while the local database of the FL client is usually small. Chawla \textit{et al.} propose an over-sampling method SMOTE~\cite{chawla2002smote}, which can generate minority classes samples to rebalance the dataset. Han \textit{et al.} improve SMOTE by considering the data distribution of minority classes~\cite{han2005borderline}. However, the above method is unsuitable for FL, because the data of clients is distributed and private. Some ensemble methods, such as AdaBoost~\cite{ratsch2001soft} and Xgboost~\cite{chen2016xgboost}, can learn from misclassification and reduce bias. However these machine learning algorithms are sensitive to noise and outliers, which are common in the distributed dataset.

\subsection{Motivation}\label{sec:motivation}
\begin{figure*}[t]
	\vspace{-3mm} 
	\setlength{\belowcaptionskip}{0.01cm} 
	\setlength{\abovecaptionskip}{-1mm} 
	\centering
	\includegraphics[scale=1.0]{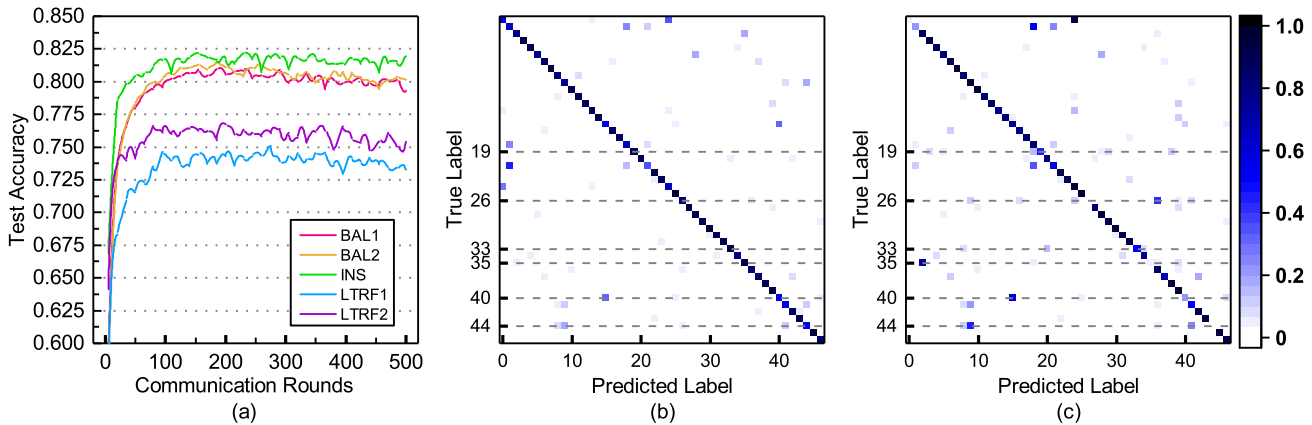}
	\caption{Accuracy and confusion matrixes on distributed EMNIST. (a) Test accuracy versus communication rounds on distributed EMNIST; (b) Comparison between the confusion matrixes of CNN models trained on BAL1 dataset and (c) LTRF1 dataset.}
	\label{motivation figure}
	\vspace{-5mm}
\end{figure*}
Federated learning is designed to be widely deployed on mobile phones and IoT devices, where each device trains model using its local data. It means that the data distribution of different devices depends on their usages which are likely to be different. For example, cameras deployed in the school capture more human pictures than the cameras deployed in the wild. 
Furthermore, another kind of imbalance is class imbalance in the collection of distributed data, such as the word frequency of English literature (following Zipf's law~\cite{adamic2002zipf}). In order to distinguish between these imbalances in federated learning, we summarize above cases into three categories: 1) Size Imbalance, where the data size on each device (or client) is uneven; 2) Local Imbalance, i.e., independent and non-identically distribution (non-IID), where each device does not follow a common data distribution; 3) Global Imbalance, means that the collection of data in all devices is class imbalanced.

To clarify the impact of imbalanced training data on federated learning, we use the FL framework to train convolutional neural networks (CNNs) based on a imbalanced dataset.
However, since there is no large distributed image classification datasets, we build new distributed datasets by resampling EMNIST~\cite{cohen2017emnist} dataset. EMNIST is a image classfication dataset which contains 47 class of handwritten English letters and digits. 
Although there has a federated version called FEMNIST~\cite{caldas2018leaf}, some unaccounted imbalances contained in training set and test set.

\textbf{Imbalance dataset}. We build five distributed EMNIST datasets: BAL1, BAL2, INS, LTRF1 and LTRF2, the detail settings are shown in TABLE~\ref{tab:motivation}. BAL1 and BAL2 are both scalar balanced and global calss balanced, the difference is BAL1 is local balanced and the local distribution of BAL2 is random. INS is a scalar imbalanced dataset and the client data size is following the images uploads number of Instagram users~\cite{amirhosein_bodaghi_2017_823283}. LTRF1 and LTRF2 futher have global imbalance by making the global class distribution following the frequency of the English letters, which is obtained through a corpus of the Simple English Wikipedia(50441 articles in total). In addition, the training data size of LTRF2 is alomost twice than of LTRF1. Note that there is no identical sample between any clients and the test set is balanced.

\begin{table}[]
	\footnotesize
	\caption{\label{tab:motivation}Settings of Distributed EMNIST Dataset.}
	\vspace{-2mm}\centering
	\begin{tabular}{@{}p{7mm}p{21mm}p{20mm}p{7mm}p{6mm}@{}}
		\toprule
		& \multicolumn{3}{c}{Types of Data Distribution} & \multicolumn{1}{c}{Sample Size} \\[-1.5pt] \midrule
		Notation & Scalar & Global & Local & Train/Test \\[-1.5pt] \midrule
		BAL1 & Even & Balanced & Balanced & 117500/18800 \\[-1pt]
		BAL2 & Even & Balanced & Random & 117500/18800 \\[-1pt]
		INS & Instagram uploads & Balanced & Random & 117500/18800 \\[-1pt]
		LTRF1 & Instagram uploads & Letters frequency & Random & 117500/18800 \\[-1pt]
		LTRF2 & Instagram uploads & Letters frequency & Random & 230752/18800 \\[-1pt]\bottomrule      
	\end{tabular}
	\vspace{-5mm}
\end{table}

\textbf{Model architecture}. The implemented CNN model has three convolution layers and two dense layers: the first two convolution layers have 12 and 18 channels, $5\times5$ and $3\times3$ kernel size (strides is 2), respectively. Above covolution layers followed by a dropout~\cite{wan2013regularization} with keep probability 0.5; The third convolution layer has 24 channels, $2\times2$ kernel size (strides is 1) and following a flatten operation. The last two dense layers are a fully connected layer with 150 units activated by ReLu and a softmax output layer. By the way, the loss function is categorical cross-entropy and the metric is top-1 accuracy. This CNN model has total 68,873 parameters and can achieve 87.85\% test accuracy after 20 epochs on EMNIST. 

\textbf{FL settings}. We use the same notation for federated learning settings as~\cite{mcmahan2016communication}: the size $B$ of local mini batch is 20 and the local epochs $E$ is 10. The total number of clients $K$ is 500 and the fraction C of clients that performs computation on each round is 0.05. 
For local training, each client updates the weights via Adam~\cite{kingma2014adam} optimizer with learning rate $\eta=0.001$ and no weight decay.

The test top-1 accuracy on five distributed EMNIST is shown in Fig.~\ref{motivation figure}(a). Experimental results show that the global imbalance leads to a significant decrease in accuracy. Qualitatively, the accuracy on BAL1 and BAL2 is 79.99\% and 80.13\%, respectively. For the global imbalanced dataset LTRF1, 7.92\% reduction in accuracy compared to INS1 was observed (from 81.60\% to 73.68\%). For LTRF2, 6.20\% reduction in accuracy compared to INS1 was observed (from 81.60\% to 75.40\%) although LTRF2 has twice amount of training data than LTRF1. 
In addition, the random local imbalance does not lead to accuracy degradation and the test accuracy is slightly improved  in scalar imbalance case (from 79.99\% to 81.60\%).


In order to elucidate the influence of global imbalance on the model, Fig.~\ref{motivation figure}(b) and Fig.~ \ref{motivation figure}(c) show the confusion matrixes of BAL1 and LTRF1. The meaning of labels is the same as EMNIST, labels 0 to 9 correspond to digitals, labels 10 to 46 corresponding English letters (15 letters are merged according to ~\cite{cohen2017emnist}). 
As shown in the confusion matrix of BAL1, most images are classified correctly, which is represented in the confusion matrix as most of the blue squares spread over the diagonal. However, for the confusion matrix of LTRF1, 6 classes of images (which correspond to the 6 letters with the lowest frequency in English writings) are not well classified as shown by the gray lines. Due to the global imbalanced training set, the CNN models are more biased towards classifying the majority classes samples.

In summary, global imbalance will cause an accuracy loss of the model trained through FL. The main challenge of the mobile FL applications is to train neural networks in the various distributed data distribution. Note that uploading or sharing users' local data is not optional because it exposes user data to privacy risks. To address the challenge, we put forward a self-balancing federated learning framework named Astraea, which improving classification accuracy by global data distribution based data augmentation and mediator based multi-client rescheduling.

\section{Design of Astraea} \label{sec:overview}

As aforementioned, the precision of federated learning on the distributed imbalanced dataset is lower than that on the balanced dataset. To find out the cause of the decline in the accuracy, we mathematically prove that the imbalance of distributed training data can lead to a decrease in accuracy of FL applications. Based on this conclusion, we design the Astraea framework, the goal of which is to relieve the global imbalance and local imbalance of clients data, and recover the accuracy.

\subsection{Mathematical Demonstration}
We define the problem of federated learning training on imbalanced dataset that leads to precision degradation. 
To show the accuracy degradation in federated learning, we use traditional SGD-based deep learning~\cite{lecun2015deep} as the ideal case and derive the update formula of optimal weights.
For SGD-based deep learning, the optimization objective is:
\begin{equation}
\min\limits_{{\bm{w}}} \mathbb{E}_{(\bm{x},\bm{y})\sim\hat{p}_{data}}L[f(\bm{x};\bm{w}),\bm{y})],
\end{equation}
where $L$ and $\hat{p}_{data}$ are the loss function and the distribution of training data, respectively. 
Since the goal is to minimize the test loss, we assume $\hat{p}_{data}={p}_{test}$, where ${p}_{test}$ means the distribution of test set that is balanced for image classification tasks. 
Both SGD-based deep learning and federated learning use the same test set. 
We assume that the initial weights for SGD-based deep learning and federated learning are the same:
\begin{equation}
\vspace{-1mm}
w_0^{(k)}=w_0=w_0^*.
\end{equation}

The optimal weights of SGD-based deep learning is updated by:
\begin{equation}
\vspace{-1mm}
\begin{split}
\bm{w}_{t+1}^*=\bm{w}_{t}^*-\eta\nabla_{\bm{w}_t^*}\sum\limits_{i=1}^nL(f(x^{(i)};\bm{w}_{t}^*),y^{(i)}),\\ (x^{(i)},y^{(i)})\sim{p}_{test}.
\end{split}
\end{equation}

Because $\bm{w}^*$ is the weights that achieves the best accuracy on the test set, so it is the optimal weights of federated learning too. 
For federated learning, the optimization objective is:
\begin{equation}
\min\limits_{{\bm{w}}}\mathbb{E}_{(\bm{x},\bm{y})\sim\hat{p}_{data}^{(k)}}L[f(\bm{x};\bm{w}^{(k)}),\bm{y})], k=1,2, ..., K.
\end{equation}
where $\hat{p}_{data}^{(k)}$ means the training data distribution of any client $k$. Given any client $k$, the corresponding training data distribution $\hat{p}_{data}^{(k)}$ is imbalanced for the considered federated learning. $\bm{w}^{(k)}$ is the weights of client $k$. The weights of each client $k$ that is optimized by gradient descent with learning rate $\eta$ is updated by:
\begin{equation}
\begin{split}
\bm{w}_{t+1}^{(k)}=\bm{w}_{t}^{(k)}-\eta\nabla_{\bm{w}_{t}^{(k)}}\frac{1}{n_{k}}\sum\limits_{i=1}^{n_{k}}L(f(x^{(i)};\bm{w}_{t}^{(k)}),y^{(i)}),\\
(x^{(i)},y^{(i)})\sim\hat{p}_{data}^{(k)}.
\end{split}
\end{equation}

The weights of federated learning server is calculated by the \textit{FedAvg}~\cite{mcmahan2016communication} algorithm:
\begin{equation}
\label{fedavg2}
\begin{split}
\bm{w}_{t+1}^{(Avg)}=&\sum\limits_{k=1}^{K}\frac{n_k}{n}\bm{w}_{t+1}^{(k)},\\
=&\sum\limits_{k=1}^{K}\frac{n_k}{n}\bm{w}_{t}^{(k)}-\frac{\eta}{n}\nabla_{\bm{w}_{t}^{(k)}}\sum\limits_{i=1}^{n_k}L(f(x^{(i)};\bm{w}_{t}^{(k)}),y^{(i)}),\\
&(x^{(i)},y^{(i)})\sim\hat{p}_{data}^{(k)}.
\end{split}
\end{equation}

Since $\hat{p}_{data}^{(k)} \neq p_{test}$, we have $\bm{w}_{t+1}^{(Avg)} \neq \bm{w}_{t+1}^*$, which means that federated learning cannot achieve optimal weights when training data distribution is imbalanced.

Next, we prove that federated learning can restore the accuracy of models if condition $\hat{p}_{data}^{(k)}={p}_{test}$ is satisfied by mathematical induction.

\textit{Proposition: }\ $\bm{w}_{t}^{(Avg)}=\bm{w}_{t}^*$ is true for $t$ is any non-negative integer and $\hat{p}_{data}^{(k)}={p}_{test}$.

\noindent{\bf Proof}:
\textit{Basis case}: Statement is true for $t=0$:
\begin{equation}
\bm{w}_0^{(Avg)}=\sum\limits_{k=1}^K\frac{n_k}{n}\bm{w}_0=\bm{w}_0^*.
\end{equation}

\textit{Inductive assumption}: Assume $\bm{w}_{t}^{(Avg)}=\bm{w}_{t}^*$ is true for $t=\mu$ and $\hat{p}_{data}^{(k)}={p}_{test}$, $\mu\in Z^+$.

Then, for $t=\mu+1$:
\begin{equation}
\begin{aligned}
\bm{w}_{\mu+1}^{(Avg)}=&\sum\limits_{k=1}^K\frac{n_k}{n}\bm{w}_{\mu}^*-
\frac{\eta}{n}\nabla_{\bm{w}_{\mu}^*}\sum\limits_{i=1}^{n}L(f(x^{(i)};\bm{w}_{\mu}^{*}),y^{(i)}),\\
=&\bm{w}_{\mu}^*-\eta\nabla_{\bm{w}_{\mu}^*}L(f(x^{(i)};\bm{w}_{\mu}^{*}),y^{(i)})=\bm{w}_{\mu+1}^*,\\
&\hat{p}_{data}^{(k)}={p}_{test}.
\end{aligned}
\end{equation}
Therefore, by induction, the statement is proved. $\hfill\blacksquare$ 

According to the above conclusion, the difference between the distributions of the training set and test set accounts for the accuracy degradation of federated learning. Therefore, to achieve a new partial equilibrium, we propose the Astraea framework to augment minority classes and create mediators to combine the skewed distribution of multiple clients. The details of Astraea are shown in the next section.

\subsection{Astraea Framework}
In order to solve the problem of accuracy degradation, the training data of each client should be rebalanced. An instinct method is to redistributing the clients' local data until the distribution is uniform. However, sharing data raises a privacy issue and cause high communication overhead. Another way to rebalance training is to update the global model asynchronously. Each client calculates updates based on the latest global model and applies its updates to the global model sequentially. It means that the communication overhead and time consumption of the method is $K$ times that of the federated learning (FL). Combining the above two ideas, we propose Astraea, which introduces mediators between the FL server and clients to rebalance training.

\begin{figure}[htbp]
	\vspace{-5mm} 
	\setlength{\belowcaptionskip}{0.01cm} 
	\setlength{\abovecaptionskip}{-2mm} 
	\centering
	\includegraphics[scale=1]{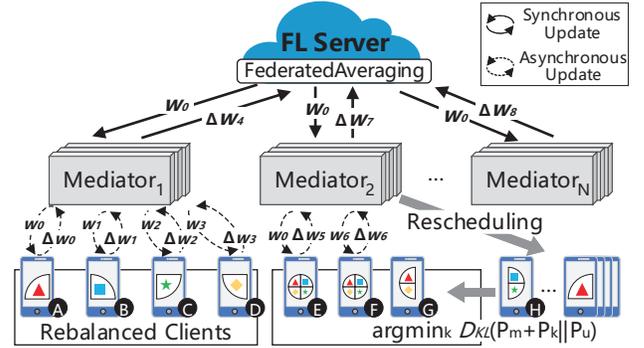}
	\caption{Astraea Framework Overview.}
	\label{framework}
	\vspace{-3mm} 
\end{figure}

\begin{figure*}[htbp]
	\vspace{-3mm} 
	\setlength{\belowcaptionskip}{0.01cm} 
	\setlength{\abovecaptionskip}{-1mm} 
	\centering
	\includegraphics[scale=1]{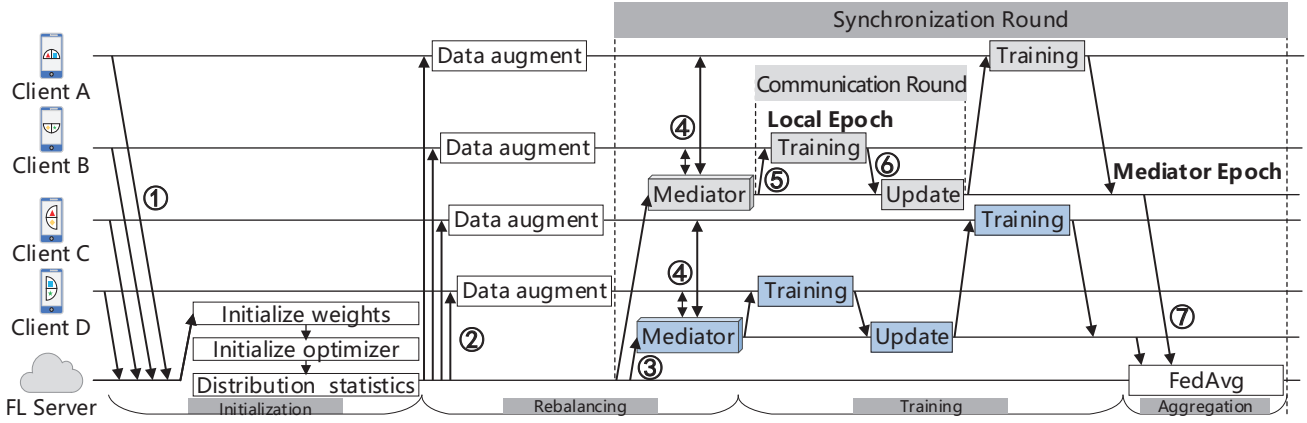}
	\caption{Astraea Workflow. Rebalance the training by data augmentation(\ding{193}) and mediator based rescheduling(\ding{194}\ding{195}).}
	\label{workflow}
	\vspace{-3mm}
\end{figure*}
The overview of the proposed Astraea framework is shown in Fig.~\ref{framework}. Astraea consists three parts: FL server, mediator, and clients. The FL server is responsible for maintaining a global model $\bm{w}_0$, deploying the model to mediators, and synchronously aggregating the updates $\Delta\bm{w}_4$, $\Delta\bm{w}_7$, $\Delta\bm{w}_8$ (as shown in Fig.~\ref{framework}) from them using the federated averaging algorithm. The clients can be mobile phones or IoT devices that maintain a local training dataset. The four shapes in the clients represent four classes of data. The clients can be divided into three categories according to their data distribution: 
\begin{itemize}
    \item Uniform clients, which have enough balanced training data and are ready to run FL applications (e.g. client E and F in Fig.~\ref{framework}).
    \item Slight clients, which have relatively small amounts of data and are hard to participate in the training process.
    \item Biased clients, which have enough training data but prefer to hold certain classes of data which leads to a global imbalance (i.e. client A-D, G, H). 
\end{itemize}
In short, the slight clients and biased clients introduce scalar imbalance and global imbalance respectively. 

Mediators have two jobs. One is to reschedule the training processing of the three kinds of clients. For example, as shown in Fig.~\ref{framework}, client G has data with label 0 and label 1, meanwhile client H has data with label 2 and label 3. Then, the mediator can combine the training of G and H to archive a partial equilibrium.

Mediators also need to make the distribution of the collection of data close to the uniform. To measure the extent of partial equilibrium, we using Kullback–Leibler divergence between $P_m+P_k$ and $P_u$, where $P_m$, $P_k$, $P_u$ means the probability distributions of mediator, rescheduling client, and uniform distribution, respectively. In addition, by combining multi-client training, a mediator can expand the size of the training set and learn more patterns than a separate client. Note that the mediators are virtual components, it can be deployed directly on the FL server or the mobile edge computing (MEC) server to reduce communication overhead. 

\begin{algorithm}[t]
	\caption{Astraea distributed neural network training.}
	\label{training}
	\small
	\begin{algorithmic}[1]
		\Procedure{FL Server training}{}
		\State \textit{Initialize $\bm{w}_0$}, $\bm{w}_1 \leftarrow \bm{w}_0$.
		\For{each synchronization round $r=1,2,...,R$}
		\For{each mediator $m$ in ${1,2,...,M}$ parallelly}
		\State $\Delta \bm{w}_{r+1}^m\leftarrow$\textbf{MediatorUpdate}($m,\bm{w}_r$).
		\EndFor
		\State $\bm{w}_{r+1} \leftarrow \bm{w}_{r}-\sum_{m=1}^{M}\frac{n_m}{n}\Delta \bm{w}_{r+1}^m$. // \textit{FedAvg}.
		\EndFor
		\EndProcedure
		
		\Function{MediatorUpdate}{$m,\bm{w}$}
		\State $\bm{w}^* \leftarrow \bm{w}$.
		\For{each mediator epoch $e_m=1,2,...,E_m$}
		\For{each clients $i$ in mediator $1,2,...,M$}
		\For{each local epoch $e=1,2,...,E$}
		\State // \textit{Asynchronous SGD}.
		\State $\bm{w}_{e}\leftarrow \bm{w}-\eta\nabla\ell(\bm{w};\mathbb{X}^{(i)},\mathbb{Y}^{(i)})$.
		\State $\bm{w} \leftarrow \bm{w}_e$.
		\EndFor
		\EndFor
		\EndFor
		\State $\Delta\bm{w} \leftarrow \bm{w}-\bm{w^*}$.
		\State \Return $\Delta\bm{w}$
		\EndFunction
	\end{algorithmic}
\end{algorithm}

Algorithm~\ref{training} shows the training process of Astraea. First, the FL server needs the initialization weights as the global model to start the training. 
Then, the FL server starts a new round $r$ of training and sends the global model to the mediators. 
Next, each mediator $m$ coordinates the assigned clients for training and calculates the updates of weights $\Delta \bm{w}_{r+1}^m$ in parallel. 
Finally, the FL server collects the updates of all mediators, aggregates the updates with the weight of ${n_m}/{n}$ ($n_m$ is the total train size for the clients assigned to mediator $m$), then updates the global model to $\bm{w}_{r+1}$ and ends this round. $\bm{w}_{r+1}$ is the start model for the next round.


\subsection{Astraea Workflow}
The workflow of Astraea includes initialization, rebalancing, training, and aggregation, as shown in Fig.~\ref{workflow}.

\textbf{Initialization}. In the initialization phase, the FL server first waits for the mobile devices to join the FL model training task. The devices participate in the training by sending their local data distribution information to the server (\ding{192}). After determining the devices (clients) to be involved in the training, the FL server counts the global data distribution and initializes the weights and the optimizer of the learning model.

\textbf{Rebalancing}. In the rebalancing phase, the server first calculates the amount of augmentations for each class based on the global data distribution. Then, all clients perform data augment in parallel according to the calculation results (\ding{193}). Algorithm~\ref{data augmentation} shows the detail of rebalancing. The \textit{Augment} function in line 11 takes one sample and generates augmentations, including random shift, random rotation, random shear, and random zoom, for the sample. The number of augmentations depends on the second parameter $(\bar{C}/C_y)^\alpha$, where $\alpha$ indicates the degree of data augmentation. Larger $\alpha$ means greater amount of augmentations. In addition, we do not augment the samples, the total sample size of their classes is greater than $\bar{C}$. The goal of data augmentation is to mitigate global imbalance rather than eliminate it while a large $\alpha$ will generate too many similar samples, which makes the model training more prone to overfitting.

\begin{algorithm}[htbp]
	\caption{Gobal data distribution based data augmentation.}
	\label{data augmentation}
	\small
	\begin{algorithmic}[1]
		\State \textbf{FL Server:}
		\State Calculate the data size of each class $C_1,..., C_N$, and the mean $\bar{C}$.
		\For {each class $i$ in $1,...,N$}
		\If {$C_i<\bar{C}$}
		\State Augmentaion set $\mathbb{Y}_{aug}\cup i$.
		\EndIf
		\EndFor
		\State
		\State \textbf{Clients:}
		\For{each client $1,...,k$ in $K$ parallelly}
		\For {each sample $(x,y)$ in client $k$ dataset $(\mathbb{X}^{(k)},\mathbb{Y}^{(k)})$}
		\If {label $y$ in augmentaion set $\mathbb{Y}_{aug}$ }
		\State $(\mathbb{X}_{aug}^{(k)},\mathbb{Y}_{aug}^{(k)})\cup$\textbf{Augment($(x,y)$, $(\bar{C}/C_y)^\alpha$)}.
		\EndIf
		\EndFor
		\State $(\mathbb{X}^{(k)},\mathbb{Y}^{(k)}) \cup (\mathbb{X}_{aug}^{(k)},\mathbb{Y}_{aug}^{(k)})$.
		\State \textbf{ShuffleDataset$(\mathbb{X}^{(k)},\mathbb{Y}^{(k)})$}.
		\EndFor
	\end{algorithmic}
\end{algorithm}

Once all the clients have completed data augmentation, the FL server creates mediators (\ding{194}) to rescheduling clients (\ding{195}) in order to achieve partial equilibrium. 
In order to get more balanced training, we can increase the collaborating clients of each mediator. However, this will also induce high communication overhead. Thus, we require that each mediator can only coordinate training for $\gamma$ clients. We will evaluate the communication overhead of mediators in Section \ref{sec:overhead}. 

The policy of rescheduling is shown in Algorithm~\ref{rescheduling}. We design a greedy strategy to assign clients to the mediators. A mediator traverses the data distribution of all the unassigned clients and selects the clients whose data distributions can make the mediator's data distribution to be closest to the uniform distribution. As shown in line 7 of Algorithm~\ref{rescheduling}, we minimize the KLD between mediator's data distribution $P_m$ and uniform distribution $P_u$. The FL server will create a new mediator when a mediator reaches the max assigned clients limitation and repeat the above process until all clients training are rescheduled.

\begin{algorithm}
	\caption{Mediator based multi-client rescheduling. $D_{KL}$ is Kullback-Leibler divergence.}
	\label{rescheduling}
	\small
	\begin{algorithmic}[1]
		\Procedure{Rescheduling}{}
		\State \textit{Initialize:}
		\State $\mathbb{S}_{mediator} \leftarrow \emptyset$, $\mathbb{S}_{client} \leftarrow 1,...,K$.
		\Repeat
		\State Create mediator $m$.
		\For {$|\mathbb{S}_{client}|>0$ and $|m|<\gamma$}
		\State $k\leftarrow \arg\min_i D_{KL}(P_m+P_i||P_u)$, $i\in\mathbb{S}_{client}$
		\State Mediator $m$ add client $k$.
		\State $\mathbb{S}_{client}\leftarrow \mathbb{S}_{client}-k$.
		\EndFor
		\State $\mathbb{S}_{mediator}\leftarrow \mathbb{S}_{mediator}\cup m$.
		\Until{$\mathbb{S}_{client}$ is $\emptyset$}
		\State\Return $\mathbb{S}_{mediator}$
		\EndProcedure
	\end{algorithmic}
\end{algorithm}

\textbf{Training}. At the beginning of each communication round, each mediator sends the model to the subordinate clients (\ding{196}). Each client trains the model with the mini-batch SGD for $E$ local epochs and returns the updated model to the corresponding mediator. The local epoch $E$ affects only the time spent on training per client and does not increase additional communication overhead. 

Then, the mediator receives the updated model \ding{197} and sends it to the next waiting training client. We call it a \emph{mediator epoch} that all clients have completed a round of training. Astraea loops this process $E_m$ times. Then, all the mediators send the updates of models to the FL server (\ding{198}). 
There is a trade-off between communication overhead and model accuracy for the mediator epochs $E_m$ times for updating the model. We will discuss the trade-off in Section \ref{sec:overhead}.

\textbf{Aggregation}. First, the FL server aggregates all the updates using \textit{FedAvg} algorithm as shown in Equation \ref{fedavg2}. Then, the FL server sends the updated model to the mediators and starts the next synchronization round. 
The main difference between Astraea and the existing FL algorithms in the model integration phase is that Astraea can achieve partial equilibrium. As a result, the integrated model in Astraea is more balanced than that in the existing federated learning algorithms. 

\section{Evaluation}

\subsection{Experimental Setup}
We implement the proposed Astraea by modifying the TensorFlow Federated Framework (TFF) ~\cite{abadi2016tensorflow} and evaluate it through the single-machine simulation runtime provided by TFF.
The notations used in the experiments are listed in the TABLE~\ref{notations}.

\textbf{Datasets and models}. We adopt two widely used datasets and the corresponding models in the evaluation: 1) Imbalanced EMNIST and its corresponding model. Same as LTRF2 dataset and the CNN model mentioned in Section \ref{sec:motivation}, $K$ and $B$ are set to 500 and 20, respectively. 2) Imbalanced CINIC-10~\cite{darlow2018cinic} and the CIFAR-10~\cite{krizhevsky2009learning} model described in Keras documentation, where $K$ and $B$ are set to 100 and 50, respectively. 
We get the imbalanced CINIC-10 by re-sampling CINIC-10 and make its global distribution following the standard normal distribution.

\textbf{Baseline}. We choose the state-of-the-art federated learning algorithm \textit{FedAvg} as the baseline ~\cite{mcmahan2016communication}, which has been applied to Google keyboard for improving query suggestions~\cite{yang2018applied}. 

\begin{table}[]
	\footnotesize
	\caption{Notations and definitions.}
	\label{notations}
	\vspace{-2mm}\centering
	\begin{tabular}{cp{7cm}}	
		\hline
		Notation& \multicolumn{1}{l}{Definition} \\ \hline
		$K$& Total number of clients. \\
		$B$& Local batch size. \\
		$c$& Number of online clients per synchronization round (per communication round if no mediator). \\
		$\alpha$& Data augmentation factor. \\
		$\gamma$& Maximum number of clients assigned to a mediator. \\
		$E_m$& Mediator epochs. All clients in a mediator are updated sequentially of $E_m$ times in a synchronization round. \\
		$E$& Local epochs. Each client updates weights $E$ times on local data in a communication round. \\ \hline
	\end{tabular}
	\vspace{-5mm}
\end{table}

\subsection{Effect of Accuracy}
We use the top-1 accuracy as metrics to the evaluate CNN models. 
We do not use other metrics, such as recall rates or F1 score because our test set is balanced and all classes of data have the same cost of misclassification. 

\begin{figure}
	\vspace{-3mm} 
	\setlength{\belowcaptionskip}{0.01cm} 
	\setlength{\abovecaptionskip}{-2mm} 
	\centering
	\includegraphics[scale=1]{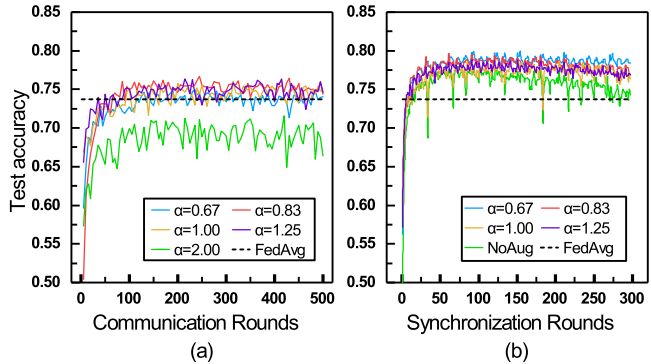}
	\caption{Comparison of accuracy on imbalanced EMNIST, $c=50,\gamma=10$. (a) Only the data augmentation strategy applied, $E=5$; (b) Combining data augmentation strategy and mediator rescheduling strategy, $E_m=2$.}
	\label{acc1}
	\vspace{-2mm}
\end{figure}

\textbf{Augmentation vs. mediator}: Fig.~\ref{acc1} shows the accuracy improvement on imbalanced EMNIST ($c=50$, $\gamma=10$), including the improved accuracy with the augmentation strategy and the improved accuracy with both rescheduling and augmentation. The experimental results show that our augmentation strategy can improve accuracy +1.28\% for $\alpha=0.83$ except when $\alpha=2$, a significant decrease in accuracy occurs. The amount of data after augmentation will greatly exceed the mean $\bar{C}$ when $\alpha=2$, which will introduce a new imbalance to the training set for the amount of augmentation is calculated by $(\bar{C}/C_y)^\alpha$. 
Hence, the recommended range for $\alpha$ is 0 to 1.

For our rescheduling strategy, the accuracy of the model is further improved from 73.77\% to 78.57\% when $\alpha=0.67$. In order to explore the accuracy improvement of rescheduling in detail, we measure the accuracy of the model that data augment is disabled. 
The results are expressed as \textit{NoAug} in Fig.~\ref{acc1}(b). The curve indicates that the accuracy is gradually reduced after 200 synchronization rounds.

\begin{figure}
	\vspace{-2mm}
	\setlength{\belowcaptionskip}{0.01cm} 
	\setlength{\abovecaptionskip}{-2mm} 
	\centering
	\includegraphics[scale=1]{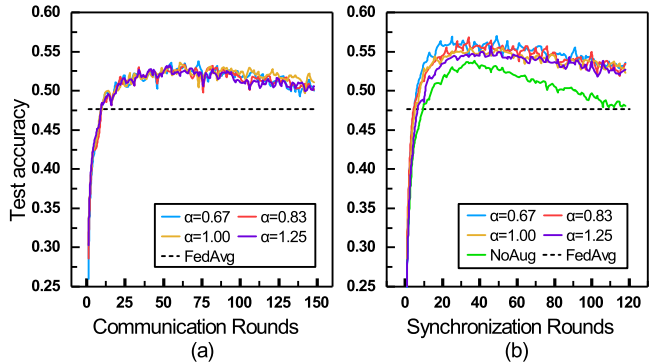}
	\caption{Comparison of accuracy on imbalanced CINIC-10, $c=50,\gamma=10$. (a) Only the data augmentation strategy is applied, $E=1$; (b) Combining data augmentation strategy and mediator rescheduling strategy, $E_m=2$.}
	\label{acc2}
	\vspace{-5mm}
\end{figure}

The accuracy improvement on imbalanced CINIC-10 is shown in Fig.~\ref{acc2}. The data augmentation strategy can improve +4.12\% top-1 accuracy when $\alpha=1.00$. The accuracy of the model is significantly improved (+5.89\% when $\alpha=0.67$) after applying the proposed rescheduling strategy. 

Similar to imbalanced EMNIST, Fig.~\ref{acc2}(b) shows that the curve of \textit{NoAug} is gradually reduced after 40 synchronization rounds. It means that the model would suffer from overfitting if augmentation is not applied. The main goal of the rescheduling strategy is to achieve partial equilibrium, which cannot solve the global imbalance. Thus, combining the two strategies is important and can achieve maximum improvement of accuracy.

\textbf{$\bm{c}$ vs. $\bm{\gamma}$}: $c$ is the number of online clients per round, which determines the scale of training in each synchronization round. $\gamma$ is the max assigned clients limitation of the mediator, which determines the scope of partial equilibrium. We explore the impact of $c$ and $\gamma$ on the training process of Astraea. 

\begin{figure}
	\vspace{-3mm} 
	\setlength{\belowcaptionskip}{0.01cm} 
	\setlength{\abovecaptionskip}{-1mm} 
	\centering
	\includegraphics[scale=1]{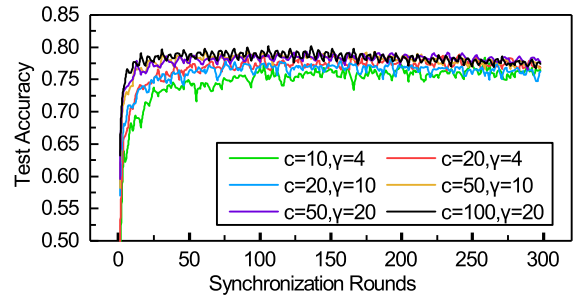}
	\caption{Test accuracy on imbalanced EMNIST, train with different numbers of participating clients per round and  different maximum assigned clients limitations of mediator.}
	\label{cgamma}
	\vspace{-3mm}
\end{figure}

The experimental results on imbalanced EMNIST are shown in Fig.~\ref{cgamma}. In the first 100 rounds, the training of model converges faster and the accuracy of the model increases with the increase of $c$. However, after 150 rounds, the accuracy is slightly reduced, especially for the models trained with a large $c$. For example, the accuracy is reduced from 79.03\% to 77.79\% when $c=100$ and $\gamma=20$. It means that the CNN models are over-training and suffered from overfitting. In order to remedy the loss of accuracy caused by overfitting, we can use the regularization strategy early stopping~\cite{yao2007early}, in which optimization is halted based on the performance on a validation set, during training. Further more, experimental results show that a larger $\gamma$ does not help improving the accuracy of the model. 

\begin{figure}
	\setlength{\belowcaptionskip}{0.01cm} 
	\setlength{\abovecaptionskip}{-1mm} 
	\centering
	\includegraphics[scale=1]{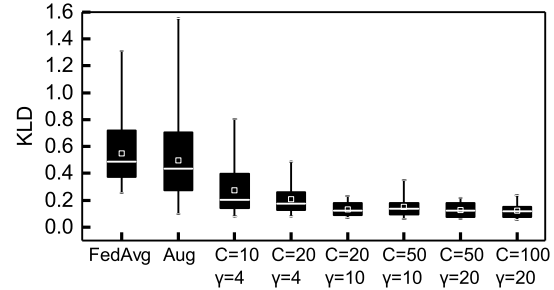}
	\caption{Kullback–Leibler divergence between $P_m$ and $P_u$ after scheduling. The horizontal axis represents different configurations of mediator, the white line indicates the mean and the white square indicates the median, the augmentation factor is 0.83.}
	\label{kld}
	\vspace{-5mm}
\end{figure}

To further explore the impact of mediator configuration on the equilibrium degree, we show the distribution of $D_{KL}(P_m||P_u)$ in Fig.~\ref{kld}. The KLD of \textit{FedAvg} is calculated by $D_{KL}(P_k||P_u)$, which means the equilibrium degree of FL framework without data augmentation and rescheduling. The KLD of $Aug$ is the equilibrium degree of Astraea framework without rescheduling. 

As shown in the Fig.~\ref{kld}, all distributions are left-skewed and the mean of KLD of \textit{FedAvg} is the highest (0.550), indicating that the distribution of \textit{FedAvg} is most imbalanced. Our augmentation strategy can make the distribution more balanced (from 0.550 to 0.498), but may introduce some new outliers. As shown in the Fig.~\ref{kld}, our rescheduling strategy can significantly rebalance data distribution (from 0.550 to 0.125) with the shrink of interquartile range and the increase of $c$. In addition, large $\gamma$ can reduce the variation of $D_{KL}$. This suggests that mediators can achieve better partial equilibrium when more clients participate in training or more clients are assigned to the mediators. In summary, the accuracy improvement of Astraea increases as the scale of the training expands.

\begin{figure}
	\vspace{-3mm} 
	\setlength{\belowcaptionskip}{0.01cm} 
	\setlength{\abovecaptionskip}{-3mm} 
	\centering
	\includegraphics[scale=1]{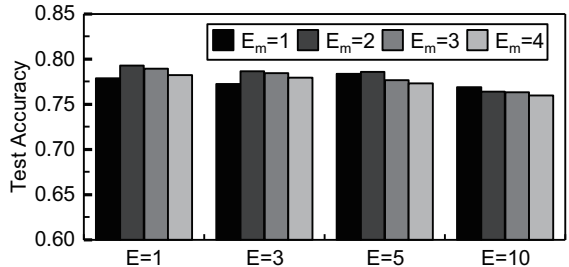}
	\caption{Test accuracy on imbalanced EMNIST. Trained with different mediator epochs $E_m$ and local epochs $E$.}
	\label{epochs}
	\vspace{-5mm}
\end{figure}

\textbf{Local epochs vs. mediator epochs}. Here we explore the impact of local epochs $E$ and mediator epochs $E_m$ on training, which represent the number of epochs for local gradients update in a communication round and the number of epochs for mediator weights update in a synchronization round, respectively. The experimental results are shown in Fig.~\ref{epochs}. 

Fig.~\ref{epochs} shows that increasing local epochs does not bring significant improvement of accuracy. A large local epochs can even cause a drop in accuracy. In our experiments, the accuracy of the CNN model drops 2.17\% on average if the local epochs of training set from 10 to 1. 
For mediator epochs, we observe that training with $E_m=2$ can significant improve accuracy (+1.4\%) compared with $E_m=1$ when the local epoch $E$ is 1.
However, the improvement achieved by mediator epochs requires additional communication resources and training time.

\subsection{Overhead}\label{sec:overhead}
We discuss three kinds of overheads of Astraea framework: Time, storage and communication. 
We ignore the computational overhead of Astraea for the additional calculations, such as augmentation and rescheduling, require few computational resources and can be calculated on the FL server. We use imbalanced EMNIST dataset in this section.

\textbf{Time overhead}. There are three major tasks that require additional time in Astraea: Data augmentation, rescheduling, and extra training epochs of the mediators. As shown in Algorithm~\ref{data augmentation}, the time complexity is $O(x^\alpha n_i)$, where $i=\arg\max_k \bar{C}/C_k$, $x=\bar{C}/C_i$. Since data augmentation is only performed once at the initialization phase, the time consumption is negligible to the whole training process. 
The process of rescheduling is shown in Algorithm~\ref{rescheduling}, where we use a greedy strategy to search the clients for rescheduling. The time complexity of the searching process is $O(c^2)$. If the data distribution of clients is static, Astraea only performs rescheduling once. In contrast, if the data distribution of the client is dynamically and rapidly changing, Astraea needs to reschedule in each synchronization round. The main time overhead of Astraea framework is the model training. In FL, the time spent on each communication round is $E\times T$, $T$ is the training time of a local epoch. In Astraea, the time spent on each synchronization round is $E_m\gamma E\times T$. 

\begin{figure}
	\vspace{-3mm} 
	\setlength{\belowcaptionskip}{0.01cm} 
	\setlength{\abovecaptionskip}{-4mm} 
	\centering
	\includegraphics[scale=1]{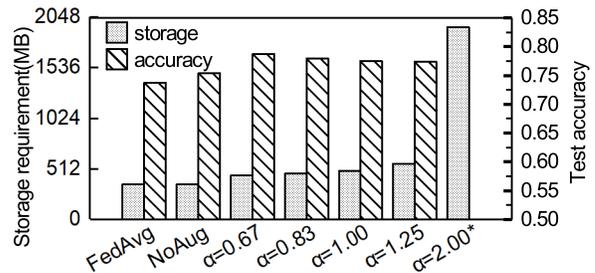}
	\caption{Overhead of storage versus improvement of accuracy on imbalanced EMNIST. The training of $\alpha=2.00$ is fail due to timeout.}
	\label{storage_overhead}
	\vspace{-3mm}
\end{figure}

\textbf{Storage overhead}. The proposed Astraea requires the clients to provide additional storage space to store the augmentation data. We show the trade-off between storage and accuracy in Fig.~\ref{storage_overhead}. The experimental results show that Astraea can improve 1.61\% accuracy on imbalanced EMNIST without additional storage requirement. It further improves 3.28\% accuracy with 25.5\% additional storage space. Although it seems that the storage overhead is large, but it is acceptable when the overhead is allotted to every client. In our experiments, the total additional storage space for data augmentation is 90 MB, i.e., 185 KB per client. The required storage space is increased with the increase of $\alpha$. $\alpha=2$ fails the training due to timeout.

\textbf{Communication overhead}. Due to the training of the clients in each mediator is asynchronous, each synchronization round in Astraea costs more traffic than each communication round in FL. The traffic of each communication round can be calculated by $2c|\bm{w}|$, where $|\bm{w}|$ is the size of all parameters. 
Hence, the traffic of each synchronization round is $2|\bm{w}|(\lceil c/\gamma \rceil +c)$. 

Experimental results in TABLE~\ref{communication_overhead2} show that Astraea is actually more communication-efficient than FL.
It is because Astraea requires less communication costs that FL in achieving a required accuracy. The communication consumption of training a CNN using FL to reach 75\% top-1 accuracy is 1176 MB whereas Astraea uses merely 215 MB (note as Med2 in TABLE~\ref{communication_overhead2}). That is, Astraea achieves 81.7\% reduction in communication cost. Although the model trained by FL can reach 75\% accuracy, it is finally stabled at around 74\%.

\begin{table}[]
	\footnotesize
	\caption{Communication consumption to reach a target accuracy for Astraea (with different value of mediator epochs note as Med1-4, $E=1$), versus FedAvg (Baseline, $E=20$), the local epochs does not affect communication overhead.}
	\label{communication_overhead2}
	\vspace{-1mm}
	\centering
	\begin{tabular}{lcccccc}
		\hline
		\multicolumn{7}{c}{\textbf{Imbalanced EMNIST}, Target Top-1 Accuracy: 75\%} \\
		Notaion& $c$ & $\gamma$ & $\alpha$ & $E$ & $E_m$ &Cost(MB)\\ \hline
		FedAvg(baseline)& 10 & \ding{55} & \ding{55} & 20 & \ding{55} & 1176 \\
		Med1& 50 & 10 & 0.67 & 1 & 1 & 284 (0.24$\times$)\\
		Med2& 50 & 10 & 0.67 & 1 & 2 & 215 (0.18$\times$)\\
		Med3& 50 & 10 & 0.67 & 1 & 3 & 221 (0.19$\times$)\\
		Med4& 50 & 10 & 0.67 & 1 & 4 & 284 (0.24$\times$)\\ \hline
	\end{tabular}
\vspace{-5mm}
\end{table}

\section{Conclusion}
Federated learning is a promising distributed machine learning framework with the advantage of privacy-preserving. However, FL does not handle imbalanced datasets well. In this work, we have explored the impact of imbalanced training data on the FL and 7.92\% accuracy loss on imbalanced EMNIST caused by global imbalance be observed. As a solution, we propose a self-balancing FL framework Astraea which rebalances the training thought 1) Performing data augmentation to minority classes; 2) Rescheduling clients by mediators in order to achieve a partial equilibrium. Experimental results show that the top-1 accuracy improvement of Astraea is +5.59\% on imbalanced EMNIST and +5.89\% on imbalanced CINIC-10 (vs. \textit{FedAvg}). Finally, we measure the overheads of Astraea and show its communication is effective.

\section*{Acknowledgment}
We would like to thank the anonymous reviewers for their valuable feedback and improvements to this paper. This work is partially supported by grants from the National Natural Science Foundation of China (61672116, 61601067, 61802038), Chongqing High-Tech Research Key Program (cstc2019jscx-mbdx0063), the Fundamental Research Funds for the Central Universities under Grant (0214005207005, 2019CDJGFJSJ001), China Postdoctoral Science Foundation (2017M620412).

\bibliographystyle{IEEEtran}
\balance
\bibliography{references}

\end{document}